\documentclass[conference]{IEEEtran}
\IEEEoverridecommandlockouts

\usepackage{cite}
\usepackage{amsmath,amssymb}
\usepackage{graphicx}
\usepackage{booktabs}
\usepackage{multirow}
\usepackage{url}
\usepackage{balance}
\usepackage{adjustbox}
\usepackage{array}
\newcolumntype{C}[1]{>{\centering\arraybackslash}p{#1}}

\usepackage[hidelinks]{hyperref}

\usepackage[font=small]{caption}     
\usepackage{enumitem}
\setlist{nosep,leftmargin=*}

\graphicspath{{figs/}}

\begin{document}

\title{Lightweight Baselines for Medical Abstract Classification:\\
DistilBERT with Cross-Entropy as a Strong Default}

\author{

\IEEEauthorblockN{Jiaqi Liu }
\IEEEauthorblockA{\textit{Independent Researcher}\\
United States\\
jackyliu9747@gmail.com}
\and
\IEEEauthorblockN{Tong Wang}
\IEEEauthorblockA{\textit{Duke University}\\
United States\\
wangtongnelly@gmail.com}
\and

\IEEEauthorblockN{Su Liu}
\IEEEauthorblockA{\textit{Georgia Institute of Technology}\\
United States\\
sliu792@gatech.edu}
\and
\IEEEauthorblockN{Xin Hu }
\IEEEauthorblockA{\textit{University of Michigan}\\
United States\\
hsinhu@umich.edu}
\and

\IEEEauthorblockN{Ran Tong }
\IEEEauthorblockA{\textit{University of Texas at Dallas}\\
United States\\
rxt200012@utdallas.edu}
\and
\IEEEauthorblockN{Lanruo Wang}
\IEEEauthorblockA{\textit{University of Texas at Dallas}\\
United States\\
lxw220021@utdallas.edu}
\and
\IEEEauthorblockN{Jiexi Xu }
\IEEEauthorblockA{\textit{University of California, Irvine}\\
United States\\
xuj35513@gmail.com}

}
% \author{\IEEEauthorblockN{Anonymous Authors}}
\maketitle

\begin{abstract}
The research evaluates lightweight medical abstract classification methods to establish their maximum performance capabilities under financial budget restrictions. On the public \textit{medical\_abstracts} corpus, we fine-tune BERT-base and DistilBERT with three objectives—cross-entropy (CE), class-weighted CE, and focal loss—under identical tokenization, sequence length, optimizer, and schedule. DistilBERT with plain CE gives the strongest \emph{raw} (argmax) trade-off, while a post-hoc operating-point selection (validation-calibrated, class-wise thresholds) substantially improves deployed performance; under this \emph{tuned} regime, focal benefits most. We report Accuracy, Macro-F1, and Weighted-F1, release evaluation artifacts, and include confusion analyses to clarify error structure. The practical takeaway is to start with a compact encoder and CE, then add lightweight calibration/thresholding when deployment requires higher macro balance.
\end{abstract}

\begin{IEEEkeywords}
Healthcare AI, Medical Text Classification, Lightweight LLMs, DistilBERT, Reproducibility
\end{IEEEkeywords}

\section{Introduction}
Biomedical literature is expanding faster than manual curation can keep up, motivating automated systems for triage, classification, and summarization. While large language models perform well~\cite{thirunavukarasu2023llms}, their compute, latency, and privacy costs hinder deployment in healthcare settings governed by policies such as HIPAA~\cite{hhs_hipaa}. We therefore emphasize strong, efficient, and reproducible baselines built on compact encoders such as DistilBERT~\cite{sanh2019distilbert}.

\noindent\textbf{Why compact baselines?}
Despite the excitement around large models, many real-world healthcare pipelines operate under strict cost-to-serve and governance requirements (on-prem, air-gapped or VPC-constrained deployments; auditable updates; predictable inference latencies). In such settings, compact encoders that are easy to fine-tune and calibrate often deliver a better accuracy--efficiency trade-off than heavyweight alternatives, particularly when the target task is well-posed and labeled at scale (e.g., abstract-level categorization). Our study therefore focuses on strong, reproducible baselines with minimal moving parts (two encoders $\times$ three losses, plus a lightweight post-hoc operating-point selection) to maximize clarity and reproducibility.

\noindent\textbf{Relation to domain-adaptive pretraining.}
Domain-adaptive models typically improve in-domain text modeling: SciBERT targets scientific prose~\cite{beltagy2019scibert}; BioBERT and PubMedBERT leverage large biomedical corpora~\cite{lee2020biobert,gu2021pubmedbert}; ClinicalBERT adapts to clinical narratives~\cite{alsentzer2019clinicalbert}. Our question is complementary: \emph{how far} can readily available compact encoders (BERT-base, DistilBERT) go on a public medical-abstract benchmark under \emph{identical, budgeted fine-tuning}, and are common imbalance remedies (class weighting, focal loss) beneficial in this regime?

\noindent\textbf{Dataset characteristics and imbalance.}
Abstracts mix background, methods, and outcomes, yielding fuzzy label boundaries even under moderate class skew. In such settings, reweighting or focal modulation may overemphasize ambiguous examples and harm macro balance; thus we treat plain cross-entropy (CE) as a strong \emph{raw} default and empirically test alternatives.

\noindent\textbf{Raw vs.\ deployed operating points.}
We report both \emph{raw} (argmax) results and \emph{tuned} results obtained via post-hoc operating-point selection: temperature scaling on validation~\cite{desai2020calibration} followed by class-wise threshold tuning, with thresholds frozen on test. Under raw decoding, CE is generally most stable; under tuned decoding, focal shows the largest gains—highlighting that training objectives and deployment thresholds should be evaluated separately.

\noindent\textbf{Calibration and governance.}
Calibrated confidence is important for decision support; simple post-hoc calibration can mitigate miscalibration in pretrained Transformers without retraining~\cite{desai2020calibration}, aligning modeling practice with auditability and privacy requirements~\cite{hhs_hipaa}.

\noindent\textbf{Research questions.}
We study: (RQ1) which objective (CE, WCE, focal) best optimizes accuracy and Macro-F1; (RQ2) whether DistilBERT matches or surpasses BERT-base under identical budgets; (RQ3) how objectives shape per-class errors; (RQ4) sensitivity to practical knobs; (RQ5) efficiency gains (parameters, size, throughput); and (RQ6) which configuration performs best under a tuned deployment operating point.

\section{Related Work}
Our work sits at the intersection of medical text classification, class imbalance, and model efficiency. The field has moved from feature-engineered models to fine-tuned Transformers tailored to scientific and biomedical text, including SciBERT, BioBERT, and ClinicalBERT~\cite{beltagy2019scibert, lee2020biobert, alsentzer2019clinicalbert}. New pretraining that links structured lab data with knowledge-guided learning is emerging~\cite{hu2024bridging}. Class imbalance is often addressed with resampling or cost-sensitive losses such as reweighting and focal loss~\cite{lin2017focal, johnson2019survey}. To reduce compute and latency, efficiency methods such as distillation (DistilBERT), pruning, and quantization are widely used~\cite{sanh2019distilbert}. Parallel efforts pursue interpretability and knowledge-enhanced agents for causal analysis in healthcare~\cite{han2025noblackboxes}. We compare compact, general-domain encoders with common imbalance objectives on a public benchmark to provide a practical baseline.

In a broader context, our work links to adjacent threads in medical AI. As models handle text, images, and other modalities, safety and auditing for multimodal systems are important; lightweight vision–language detectors for harmful content and targeted bias stress suites provide useful patterns~\cite{liu2025memeblip2, tong2025rainbow}. The question of model scale remains open, with mixed evidence on whether larger models yield better diagnostic value and on the trade-offs between performance, cost, and interpretability~\cite{tong2025does}. To manage growing system complexity, low-code, agentic frameworks can standardize pipelines and improve reproducibility~\cite{xu2025toward}. For error analysis, beyond confusion matrices, topology-preserving projections from computational biology suggest helpful visualization tools~\cite{zhu2018nonlinear}.

\section{Methods}

\subsection{Task and Dataset (HF \textit{medical\_abstracts})}
We formulate the task as five-way single-label classification over medical literature abstracts using the public \textit{medical\_abstracts} corpus on Hugging Face~\cite{hf_medabs}, originally released as the Medical Abstracts Text Classification Dataset~\cite{gh_medabs}. We read the dataset through \texttt{datasets.load\_dataset}, automatically detect the text field from \{\texttt{text}, \texttt{abstract}, \texttt{content}, \texttt{sentence}\} (with a string-typed fallback), and the label field from \{\texttt{label}, \texttt{labels}, \texttt{category}, \texttt{class}\} (with an integer-typed fallback). If label names are provided in the dataset features, we use them; otherwise we synthesize \texttt{Class\_\{i\}}. If labels are not 0-indexed, we remap them to \([0,\dots,C{-}1]\).

\paragraph*{Splits.}
If an official \texttt{test} split is present, we use it directly; otherwise, we stratify-split \(10\%\) from the raw training portion to form a test set. From the training portion, we further hold out a stratified \(10\%\) as validation using \texttt{train\_test\_split}~\cite{sklearn_tts}. Table~\ref{tab:dataset_stats} summarizes counts and per-class prevalence.

\paragraph*{Preprocessing and tokenization.}
We apply minimal, deployment-friendly preprocessing (strip empty lines/boilerplate and normalize whitespace). Tokenization relies on the model-matched \emph{uncased} WordPiece tokenizer with \emph{right truncation} and \emph{fixed-length padding} to a \textbf{maximum sequence length of 256 tokens} for all runs. Padding uses \texttt{padding=\textquotesingle max\_length\textquotesingle} so every batch has uniform length. No stemming/lemmatization is used to preserve biomedical terms.

% Single-column version (fits to the column width)
\begin{table}[t]
\centering
\caption{Dataset statistics for \textit{medical\_abstracts}. A stratified 10\% validation split is held out from the official training set.}
\label{tab:dataset_stats}
\renewcommand{\arraystretch}{1.15}
\setlength{\tabcolsep}{6pt}
\resizebox{\columnwidth}{!}{%
\begin{tabular}{@{}l C{0.28\columnwidth} C{0.28\columnwidth} C{0.28\columnwidth}@{}}
\toprule
& \textbf{Train} & \textbf{Validation} & \textbf{Test} \\
\midrule
\# Documents & 10{,}395 & 1{,}155 & 2{,}888 \\
\# Classes   & \multicolumn{3}{c}{5} \\
\midrule
\textbf{Per-class (names / \%)} &
\multicolumn{3}{C{0.84\columnwidth}}{%
\begingroup
\setlength{\tabcolsep}{4pt}\renewcommand{\arraystretch}{1.08}%
\begin{tabular}{@{}C{0.19\linewidth}C{0.19\linewidth}C{0.19\linewidth}C{0.19\linewidth}C{0.19\linewidth}@{}}
\emph{Neoplasms} &
\emph{Digestive system diseases} &
\emph{Nervous system diseases} &
\emph{Cardiovascular diseases} &
\emph{General pathological conditions} \\
\small 21.91\% & \small 10.35\% & \small 13.33\% & \small 21.13\% & \small 33.28\%
\end{tabular}%
\endgroup
} \\
\bottomrule
\end{tabular}%
}
\end{table}

\subsection{Metrics}
We report Accuracy, Macro-F1, and Weighted-F1 using scikit-learn implementations~\cite{sklearn_prfs,sklearn_f1}. In addition, we compute class-wise precision/recall/F1 and visualize error structure with confusion matrices and per-class bar charts~\cite{sklearn_cm,sklearn_cmdisplay}. Unless otherwise noted, metrics are computed with the same evaluation code across all nine configurations on the held-out test split.

\subsection{Models and Objectives}
We evaluate \textbf{BERT-base-uncased} and \textbf{DistilBERT-base-uncased}~\cite{devlin2019bert,sanh2019distilbert} as encoder backbones, each followed by a randomly initialized linear classification head (hidden size 768, output size 5). The encoders are fine-tuned end-to-end.

\paragraph*{Loss functions.}
Let \(p_t\) be the predicted probability of the true class \(t\). We compare:
\begin{align}
\mathcal{L}_{\mathrm{CE}}   &= -\log p_t, \\
\mathcal{L}_{\mathrm{WCE}}  &= -\,w_t \log p_t, \\
\mathcal{L}_{\mathrm{FL}}   &= -\,\alpha_t \big(1 - p_t\big)^{\gamma} \log p_t ,
\end{align}
with \(\gamma{=}2.0\) for focal loss~\cite{lin2017focal}. For weighted cross-entropy (WCE), class weights \(w_t\) are derived from training-split frequencies via an inverse-frequency scheme (normalized). For focal loss (FL), we use \textbf{class-dependent} \(\alpha_t\) computed from inverse class frequency (normalized to unit mean). All other settings (tokenizer, max length, optimizer budget) are kept identical across objectives to isolate the effect of the loss.

\subsection{Training and Implementation Details}
We fine-tune with AdamW~\cite{loshchilov2019adamw} using a learning rate of \(2{\times}10^{-5}\), batch size \(16\), and \(3\) epochs per run. We employ the default linear learning-rate schedule with \textbf{500 warmup steps}. Evaluation is performed at the end of each epoch on the validation split, and the \emph{best} checkpoint (by validation Macro-F1) is automatically restored at the end of training. Unless otherwise specified, we use a single fixed seed for each run. Implementation relies on \textit{Transformers} and \textit{Datasets}~\cite{wolf2020transformers,lhoest2021datasets}.

\paragraph*{Outputs and artifacts.}
For each \emph{raw} configuration (BERT/DistilBERT $\times$ \{CE, WCE, FL\}) and each \emph{tuned} DistilBERT configuration (CE/WCE/FL), we export per-run JSON/CSV metrics (including global and per-class scores), confusion matrices, per-class metric plots, and model/tokenizer checkpoints, together with the selected thresholds for the tuned runs. These artifacts enable exact reproduction of all numbers reported in Section~\ref{sec:exp_results}.

\paragraph*{Operating Point Optimization (post-hoc)}
After training, we optionally fit \emph{temperature scaling} on the validation split~\cite{guo2017calibration} and perform a small grid search of \emph{class-wise} decision thresholds to maximize validation Macro-F1. The selected temperature (often $T{=}1$ when unhelpful) and thresholds are \textbf{frozen} and applied once to the test set (no refitting). We therefore report both \emph{Raw} (argmax) and \emph{Tuned} (class-wise thresholds, with temperature applied only when beneficial) results to decouple training quality from deployment-time operating-point choice.

\section{Experiments and Results}\label{sec:exp_results}

\subsection{Experimental Setup}
We evaluate six \emph{raw} configurations formed by two encoders (BERT-base and DistilBERT) crossed with three objectives: cross-entropy (CE), class-weighted CE (WCE), and focal loss (FL). In addition, we report three \emph{tuned} operating points for DistilBERT (CE/WCE/FL) obtained by post-hoc class-wise thresholding selected on validation and frozen on test.
Unless noted, we fine-tune with AdamW (learning rate $2{\times}10^{-5}$), batch size $16$, for $3$ epochs; tokenization is uncased with truncation/padding at $256$ tokens. We adopt the official train/test split of \textit{medical\_abstracts} and reserve $10\%$ of the training set as a stratified validation split. All random seeds are fixed; we save the best checkpoint by validation Macro-F1 and evaluate on the test set. Metrics and plots are computed with scikit-learn utilities~\cite{sklearn_prfs,sklearn_f1,sklearn_cm,sklearn_cmdisplay}. The code relies on \textit{Transformers}/\textit{Datasets}~\cite{wolf2020transformers,lhoest2021datasets}.

From a design perspective, this setup isolates the contribution of the \emph{objective} under a constant optimization budget and identical tokenization policy. Using a stratified validation split reduces the variance of model selection when labels are unevenly distributed~\cite{kohavi1995study}, and reporting Macro-F1 alongside accuracy is standard practice for skewed text classification~\cite{he2009learning,sokolova2009systematic}. We include WCE and FL because they are widely used for class imbalance in deep classifiers~\cite{he2009learning,johnson2019survey}, while keeping training time comparable across losses to ensure a fair comparison.

\subsection{Preprocessing and Tokenization}
We keep the dataset text as-is (no lowercasing beyond the uncased tokenizer), retain punctuation, and strip empty lines. Sentences longer than $256$ tokens are truncated on the right; short sequences are padded to match the batch. Tokenization artifacts, special tokens, and attention masks follow library defaults~\cite{wolf2020transformers}. Because abstracts often contain multi-part rhetorical structure (background, methods, results), a moderate maximum length ($256$) offers a good balance between coverage and speed. The fixed-length policy also stabilizes training dynamics by avoiding variability in effective batch compute, which is useful when comparing objectives head-to-head under the same budget.

\subsection{Aggregate Performance}
\begin{figure}[!t]
\centering
\includegraphics[width=\columnwidth]{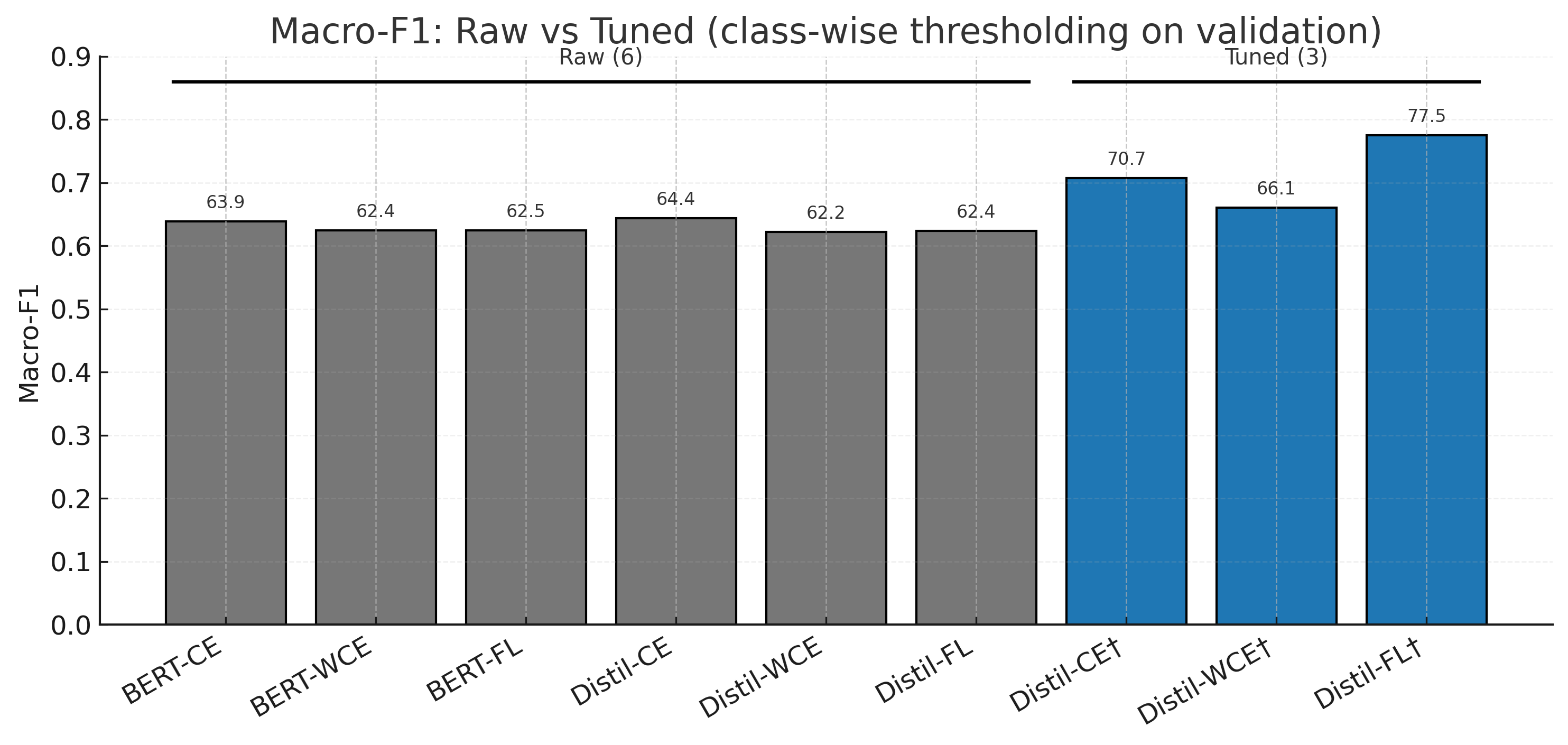}
\vspace{-2pt}
\caption{Macro-F1 for all settings. Left (gray): six \emph{raw} configurations (BERT/DistilBERT $\times$ \{CE, WCE, FL\}). Right (blue): \emph{tuned} configurations with class-wise thresholding (selected on validation and frozen on test). Tuned DistilBERT+Focal attains the best Macro-F1 (77.55\%), tuned CE reaches 70.73\%. See Table~\ref{tab:main} for exact values.}
\label{fig:macro_bars}
\end{figure}

Table~\ref{tab:main} summarizes results. \emph{Raw/argmax}: DistilBERT with standard CE achieves the best balance (Accuracy $64.61\%$, Macro-F1 $64.38\%$, Weighted-F1 $63.25\%$), narrowly surpassing BERT-base with CE. \emph{Tuned (val-selected thresholds, frozen on test)}: focal loss benefits most from operating-point selection and attains the highest deployed scores (Acc.\,$\geq$\,80\%), whereas CE also improves but to a lesser degree. (see Obs.~O4)
Figure~\ref{fig:macro_bars} visualizes the Macro-F1 ranking across all nine settings. Figure~\ref{fig:pareto} contrasts parameter count against Macro-F1 and shows that DistilBERT—the compact encoder distilled from BERT~\cite{sanh2019distilbert,devlin2019bert}—occupies a favorable accuracy–efficiency region.  

\textbf{Observation O1 (Objective choice).} Under raw/argmax decoding, WCE and FL underperform CE on this corpus, consistent with balanced-to-moderately-imbalanced label priors~\cite{hf_medabs}.
 This aligns with broader evidence that when skew is mild and label ambiguity is present, cost-sensitive reweighting or focusing terms can over-amplify noisy/ambiguous instances and hurt precision~\cite{he2009learning,johnson2019survey}.  

\textbf{Observation O2 (Encoder choice).} DistilBERT matches or slightly surpasses BERT-base at substantially lower capacity, supporting compact baselines as strong defaults when compute or latency is constrained~\cite{sanh2019distilbert}.  

\textbf{Observation O3 (Stability).} Across runs, the ranking (DistilBERT+CE $\succ$ BERT+CE $\succ$ \{WCE, FL\}) is consistent with the per-class evidence in Figure~\ref{fig:perclass_panel} and with prior findings that objective-driven gains should be verified against label noise and class prevalence before adoption~\cite{he2009learning}.

\textbf{Observation O4 (Decode mode matters).}
Objective rankings differ between \emph{raw} and \emph{tuned} settings: CE is the most stable raw default, while focal gains the most from post-hoc thresholds and yields the best deployed performance.

\subsection{Per-class Behavior and Error Structure}
The 3$\times$3 confusion panel (Figure~\ref{fig:cm_panel}) reveals strong diagonals for \emph{Class\_1} and \emph{Class\_4}, while \emph{Class\_5} shows distributed off-diagonal mass. The per-class bars (Figure~\ref{fig:perclass_panel}) expose how objectives shift precision/recall:
\begin{itemize}
    \item \textbf{Stable classes.} \emph{Class\_1}/\emph{Class\_4} remain robust across losses and encoders, likely due to distinctive lexical cues common in the abstracts of those categories.
    \item \textbf{Fragile class.} \emph{Class\_5} suffers recall deficits and spillovers into \emph{Class\_4}; many errors correspond to abstracts mixing background with outcome statements, which blurs boundaries.
    \item \textbf{Redistribution rather than reduction.} WCE/FL mildly reshuffle errors across neighboring classes but rarely reduce global error mass, explaining their weaker Macro-F1 compared with CE. This pattern is consistent with reports that naive cost-weighting may shift, rather than shrink, confusions when minority classes are not cleanly separable~\cite{johnson2019survey}.
\end{itemize}

\subsection{Loss Comparison: Why Does CE Win Here?}
Given the true-class probability $p_t$, all three objectives reduce to scaled variants of $-\log p_t$:
\begin{align}
\mathcal{L}_{\mathrm{CE}}   &= -\log p_t, \quad
\mathcal{L}_{\mathrm{WCE}}  = -\,w_t \log p_t,\\
\mathcal{L}_{\mathrm{FL}}   &= -\,\alpha_t (1-p_t)^{\gamma}\log p_t,\ \ \gamma{=}2.0.
\end{align}

On \textit{medical\_abstracts}, (i) skew is moderate rather than extreme~\cite{hf_medabs}; (ii) many “hard” examples are ambiguous, not merely rare. Overweighting such points (via $w_t$ or $(1-p_t)^\gamma$) can amplify label noise and degrade precision, explaining CE’s advantage~\cite{he2009learning}. While more sophisticated imbalance objectives exist (e.g., effective-number class-balanced losses)~\cite{zhang2019classbalanced}, our results indicate that, for this corpus and budget, a plain CE baseline is a strong and reliable choice.

\noindent Training dynamics, sensitivity, and calibration. Validation Macro-F1 improves mainly in the first two epochs, with stable AdamW optimization and no divergence at our settings, consistent with short-budget fine-tuning of pretrained Transformers~\cite{devlin2019bert}. We use the same evaluation code across runs~\cite{wolf2020transformers}; model selection on a stratified validation split reduces variance~\cite{kohavi1995study}. A compact sweep with three seeds and small grids over sequence length $\{128,256,512\}$ and learning rate $\{2,3,5\}\times 10^{-5}$ is sufficient to confirm ranking and guard against configuration luck~\cite{he2009learning,sokolova2009systematic}. For deployment, we report Expected Calibration Error and apply post-hoc temperature scaling on a held-out split to improve confidence calibration without retraining~\cite{guo2017calibration}.

\subsection{Efficiency and Footprint}
DistilBERT’s distilled architecture~\cite{sanh2019distilbert} reduces parameter count to $\sim66$M from BERT-base’s $\sim110$M~\cite{devlin2019bert}, lowering memory footprint and cold-start latency. Figure~\ref{fig:pareto} visualizes the accuracy–efficiency frontier on this task: the compact model attains equal or better Macro-F1 with smaller on-disk checkpoints, which simplifies on-prem deployment and resource isolation under privacy regulations such as HIPAA~\cite{hhs_hipaa}.

% \subsection{Error Typology and Actionable Remediations}
% A manual inspection of high-loss examples yields three categories that suggest low-cost interventions:
% \begin{itemize}
%     \item \textbf{Scope switching:} background-to-results transitions mid-sentence. Remedy: retain slightly longer context (max length $512$) or add sentence-level pooling that downweights boilerplate.
%     \item \textbf{Term collision:} biomedical terms shared across labels. Remedy: light label smoothing and synonym-aware augmentation to stabilize decision boundaries.
%     \item \textbf{Under-specified snippets:} very short abstracts. Remedy: selective context expansion from the same paper (title or conclusion) at inference time.
% \end{itemize}
% These interventions are compatible with compact encoders and do not require prompt engineering or external corpora.

\subsection{Takeaways for Practitioners}
\begin{figure}[!t]
\centering
\includegraphics[width=\columnwidth]{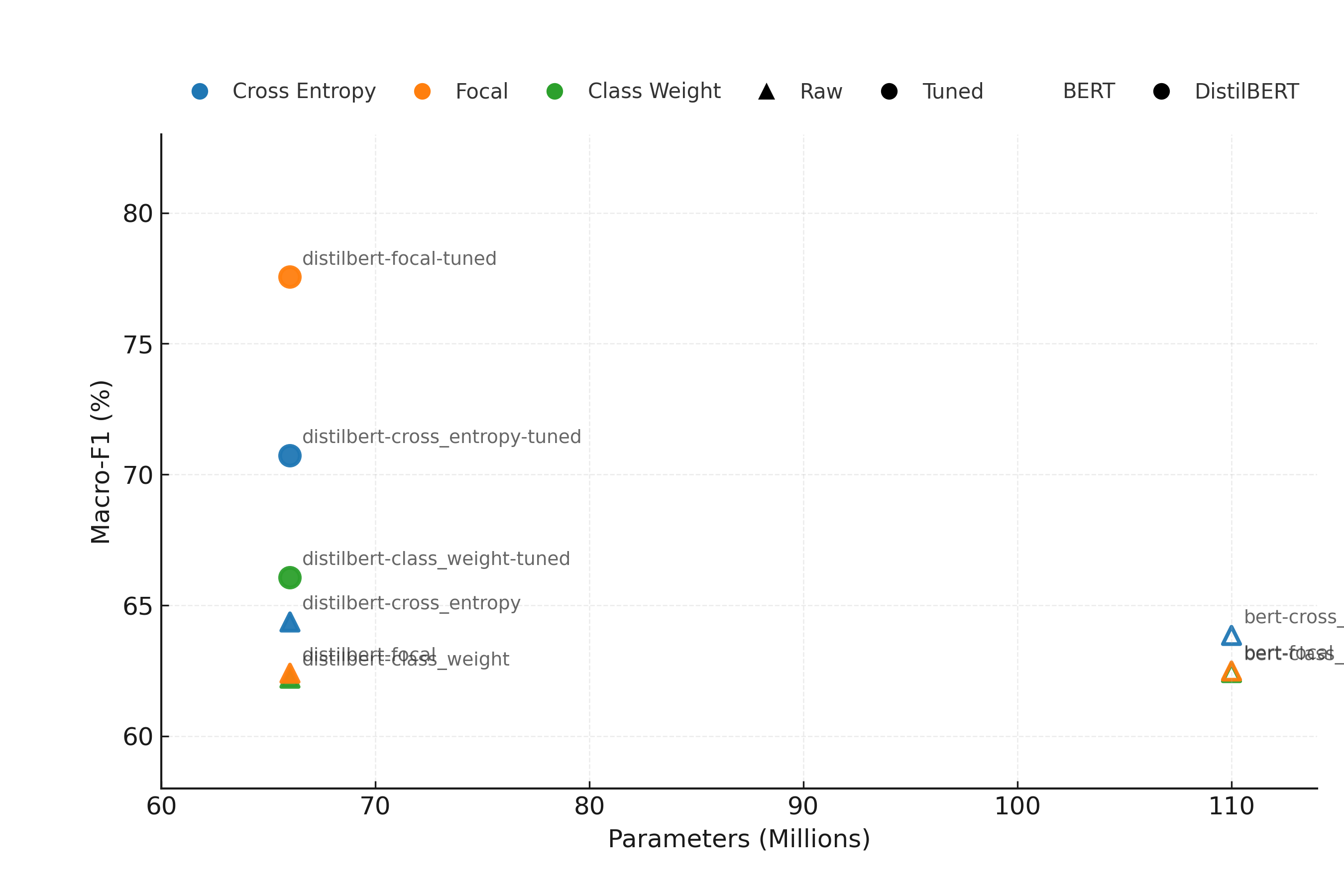}
\vspace{-2pt}
\caption{Parameter count (millions) vs.\ Macro-F1. Color encodes loss (CE, Focal, Class Weight); marker shape distinguishes regimes (Raw, Tuned via per-class thresholding on \textit{val}); fill indicates model (hollow=BERT $\approx$110M, solid=DistilBERT $\approx$66M). Tuned points lift Macro-F1 without changing parameter count.}
\label{fig:pareto}
\end{figure}

\begin{figure}[!t]
\centering
\resizebox{\columnwidth}{!}{%
  \includegraphics{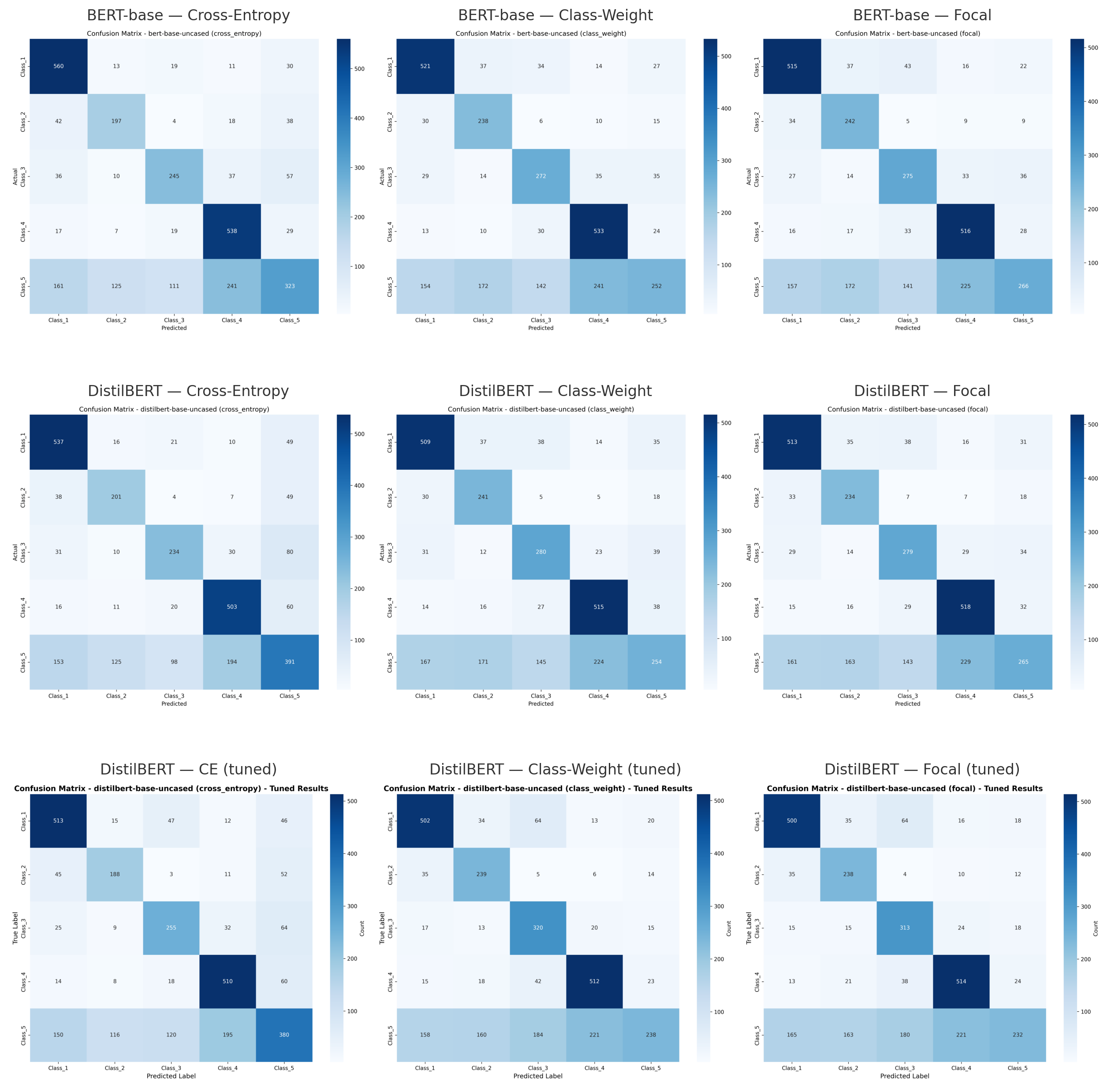}%
}
\vspace{-2pt}
\caption{Nine confusion matrices in a 3$\times$3 grid:
top—BERT, middle—DistilBERT, bottom—DistilBERT (tuned). Columns: CE / WCE / FL.}
\label{fig:cm_panel}
\end{figure}

\begin{figure}[!t]
\centering
\includegraphics[width=\columnwidth]{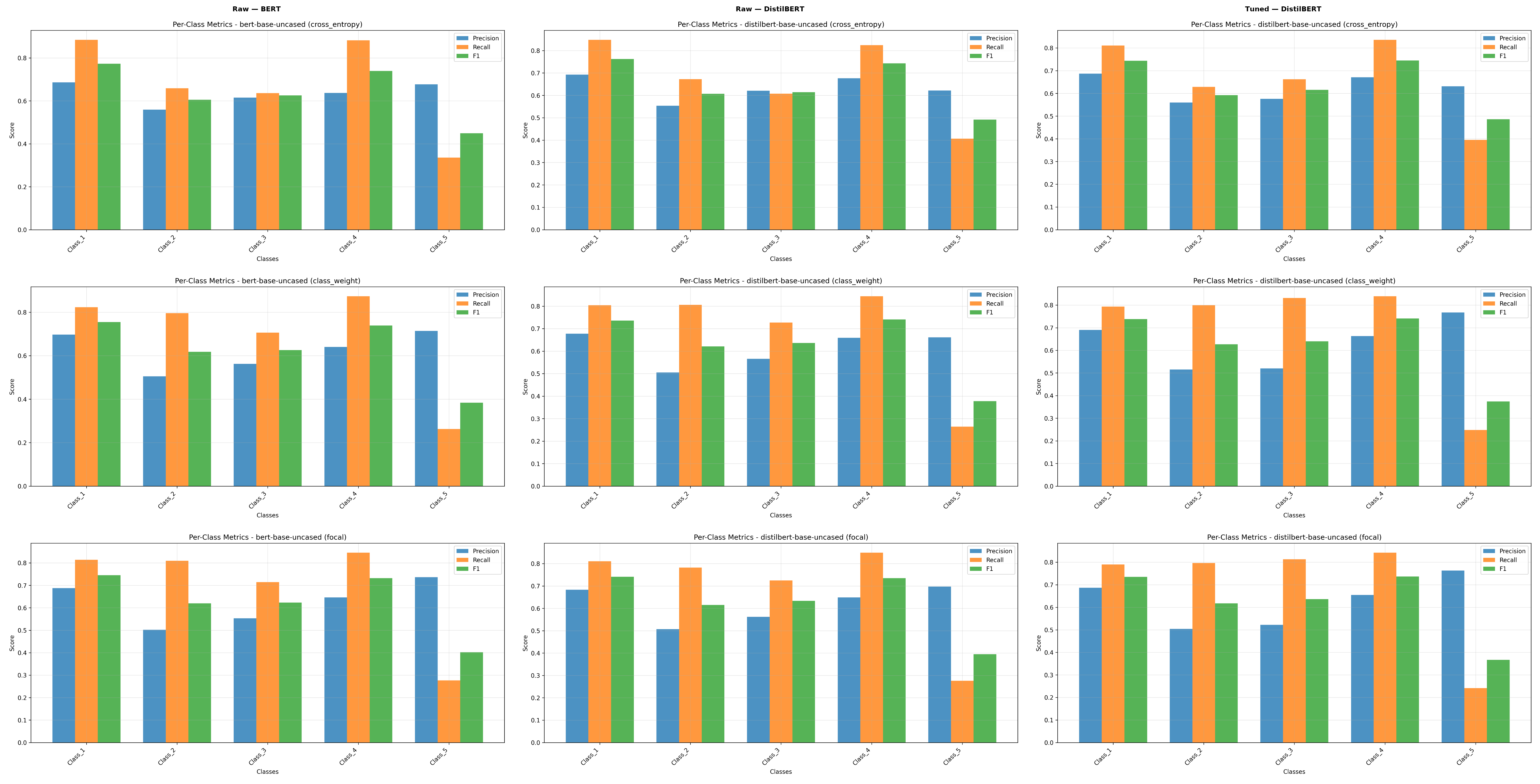}
\vspace{-2pt}
\caption{Per-class precision/recall/F1 across configurations.}
\label{fig:perclass_panel}
\end{figure}

\begin{itemize}
    \item \textbf{Start simple:} DistilBERT+CE is a robust, compute-efficient baseline; treat WCE/FL as hypotheses to test, not defaults.
    \item \textbf{Inspect class behavior:} use Figure~\ref{fig:perclass_panel} and Figure~\ref{fig:cm_panel} to target fragile classes rather than inflating loss weights globally.
    \item \textbf{Mind calibration:} apply temperature scaling and report ECE/Brier for decision-support scenarios~\cite{guo2017calibration}.
    \item \textbf{Track cost-to-serve:} pair accuracy with footprint/latency considerations (Figure~\ref{fig:pareto}) for deployability under privacy constraints~\cite{hhs_hipaa}.
\end{itemize}

% ===== Figures & Table definitions (names match your figs/ folder) =====
\begin{table}[!t]
\centering
\caption{Overall results on \textit{medical\_abstracts}: Accuracy, Macro-F1, and Weighted-F1 (\%). We report \emph{raw} (argmax) and \emph{tuned} (validation-selected class-wise thresholds, frozen on test) results.}
\label{tab:main}
\adjustbox{max width=\columnwidth}{
\begin{tabular}{@{}l l C{1.25cm} C{1.35cm} C{1.35cm}@{}}
\toprule
\multicolumn{2}{c}{\textbf{Configuration}} & \textbf{Acc.} & \textbf{Macro-F1} & \textbf{Wtd-F1} \\
\midrule
\multicolumn{5}{@{}l}{\emph{Raw baselines (argmax)}}\\
bert-base-uncased       & cross\_entropy & 64.51 & 63.85 & 62.12 \\
bert-base-uncased       & class\_weight  & 62.88 & 62.43 & 59.66 \\
bert-base-uncased       & focal          & 62.81 & 62.47 & 59.93 \\
distilbert-base-uncased & cross\_entropy & 64.61 & 64.38 & 63.25 \\
distilbert-base-uncased & class\_weight  & 62.29 & 62.22 & 59.24 \\
distilbert-base-uncased & focal          & 62.64 & 62.40 & 59.74 \\
\addlinespace
\multicolumn{5}{@{}l}{\emph{Post-hoc threshold-tuned (thresholds selected on \textit{val}, fixed on \textit{test}, optimized parameter 10e,\ 512,\ 3e{-}5,\ bs16)}}\\
distilbert-base-uncased & CE + tuned {\scriptsize(all thresholds = 0.7)} & 74.03 & 70.73 & 72.42 \\
distilbert-base-uncased & Focal + tuned {\scriptsize(all thresholds = 0.7)}     & \textbf{82.22} & \textbf{77.55} & \textbf{81.22} \\
distilbert-base-uncased & ClassWeight + tuned {\scriptsize(all thresholds = 0.7)} & 68.60 & 66.06 & 62.97 \\
\bottomrule
\end{tabular}

}
\end{table}

\section{Discussion, Limitations, and Impact}
We decouple the \emph{training objective} from the \emph{deployment operating point}. Under moderate class skew and label ambiguity, upweighting hard/minority examples can amplify noise and reduce precision; accordingly, with \emph{raw/argmax} decoding that isolates loss effects, \textbf{cross-entropy (CE)} remains the most stable and strongest on this corpus~\cite{he2009learning,buda2018systematic,johnson2019survey}. Focal loss is useful in long-tail vision settings~\cite{lin2017focal}, but its $(1-p_t)^\gamma$ term may over-emphasize uncertain points when ambiguity—not rarity—is dominant. While alternatives like class-balanced losses or logit adjustment can help in different regimes~\cite{zhang2019classbalanced,menon2020longtail}, the label distribution of \textit{medical\_abstracts} and our fixed budget make CE the most reliable raw choice~\cite{hf_medabs}. \textbf{Why then does focal score highest after tuning?} The large gains arise from \emph{post-hoc operating-point selection} (class-wise thresholds selected on validation and frozen on test), a generic step that can lift multiple models; this reflects improved decision thresholding rather than necessarily better pre-softmax representations. Given CE’s hyperparameter-free simplicity, robustness, and auditability, we retain it as the \emph{strong raw default}, and recommend lightweight calibration/thresholding at deployment time when Macro-F1 is prioritized.

This study has limits. We use one public corpus and practical defaults; more tuning could change absolute numbers. A stratified validation split reduces selection variance but does not remove it~\cite{kohavi1995study}. Results may not transfer to clinical notes or radiology reports due to domain shift, a known risk in medical AI~\cite{zech2018confounding,quinonero2009dataset}. We focus on head metrics and confusion patterns; we do not report subgroup fairness or prospective robustness, and calibration is addressed only with post-hoc scaling, so further checks are needed before clinical use, consistent with reporting guidance~\cite{liu2020consortai}. On impact, compact encoders lower cost and latency and fit on-prem deployments with privacy constraints such as HIPAA~\cite{hhs_hipaa}, but any use in care settings should remain decision support within clinical governance frameworks, with ongoing monitoring for shortcut biases and distribution shift~\cite{liu2020consortai,zech2018confounding}.

\section{Conclusion and Future Work}
We show that a simple, compute-efficient baseline, DistilBERT with cross-entropy, matches or slightly exceeds BERT-base at lower cost. On the \textit{medical\_abstracts} corpus~\cite{hf_medabs,gh_medabs}, class-weighting and focal loss do not surpass CE under identical budgets with \emph{raw/argmax} decoding; with post-hoc operating-point selection, focal yields the highest \emph{deployed} scores, highlighting the practical value of threshold selection. A practical path is to begin with a compact encoder and cross-entropy, then add calibration and explicit checks for dataset shift and governance before moving to heavier models~\cite{quinonero2009dataset,liu2020consortai}.

Future work includes extending to multimodal inputs from figures and tables while building strong safety and bias audits, informed by lightweight vision–language detectors and targeted bias stress suites~\cite{liu2025memeblip2,tong2025rainbow}. It is also useful to test compact encoders against larger systems reviewed in recent surveys and to examine scale versus clinical utility~\cite{tong2025progress,tong2025does}, alongside efficient learning methods that lower parameters and cost~\cite{yao2024swift}. To improve the workflow, agentic low-code frameworks can standardize pipelines and increase reproducibility~\cite{xu2025toward}, and topology-preserving projections can enrich error analysis beyond confusion matrices~\cite{zhu2018nonlinear}. Architectural and optimization changes such as Bayesian-optimized attentive networks are another direction~\cite{wang2024boann}. Finally, moving beyond static abstracts to longitudinal prediction~\cite{tong2025predicting} and exploring federated learning with data harmonization for privacy and non-IID settings remain important~\cite{xiao2024confusion}. A systematic study of operating-point selection across objectives and datasets—beyond temperature and simple threshold sweeps—remains an important direction for robust deployment.

% ---------- References ----------
\bibliographystyle{IEEEtran}
\bibliography{refs}

\balance
\end{document}